\documentclass[runningheads]{llncs}
\usepackage[T1]{fontenc}
\usepackage{graphicx}

\usepackage{tcolorbox}
\usepackage{listings}
\usepackage{iftex}
\ifLuaTeX
  \usepackage{luatexja}
\else
  \usepackage{xeCJK}
  \setCJKmonofont{HaranoAjiMincho-Regular.otf}
\fi
\usepackage{fancyvrb}
\usepackage{amssymb}
\usepackage{bbding}

\begin{document}

\title{Structured Maps for Landmark-Based \\Route Instructions with LLMs}

\author{
Mifu~Matsumoto\inst{1}\Envelope\orcidID{0009-0001-5392-8393} \and \\
Yuki~Nakanishi\inst{1}\orcidID{0009-0006-3226-8564} \and \\
Tomoya~Hashiguchi\inst{1}\orcidID{0000-0003-0424-2755} \and \\
Hiroaki~Ohshima\inst{1}\orcidID{0000-0002-9492-2246}}
\authorrunning{M. Matsumoto et al.}

\institute{
University of Hyogo, Hyogo, Japan
}
\maketitle

\begin{abstract}
To generate landmark-based route instructions, models must understand both map geometry and human landmark preferences. 
Although large language models (LLMs) show great abilities in language tasks, the best way to structure map data for these models is not yet established.
This paper proposes a structured text representation for landmark selection tasks in route navigation, utilizing SVG format to encode spatial features.
The structured text maintains a geometric connection with rendered map images, allowing language models to directly access spatial and semantic data. 
Using a new dataset created from human-written route descriptions and OpenStreetMap data, we evaluate models on two tasks: landmark selection at turns and around destinations. 
Experiments with five LLMs show that the proposed structured text input performs as well as or better than image-based PNG inputs. 
In addition, task-specific fine-tuning enables smaller local LLMs to effectively leverage map layouts to achieve high selection accuracy. 
These results demonstrate that structured text is an efficient and lightweight interface for spatial reasoning in human-oriented navigation.
\end{abstract}

\section{Introduction}
\label{sec:introduction}
\begin{figure}[t]
  \centering
  \includegraphics[width=0.85\linewidth]{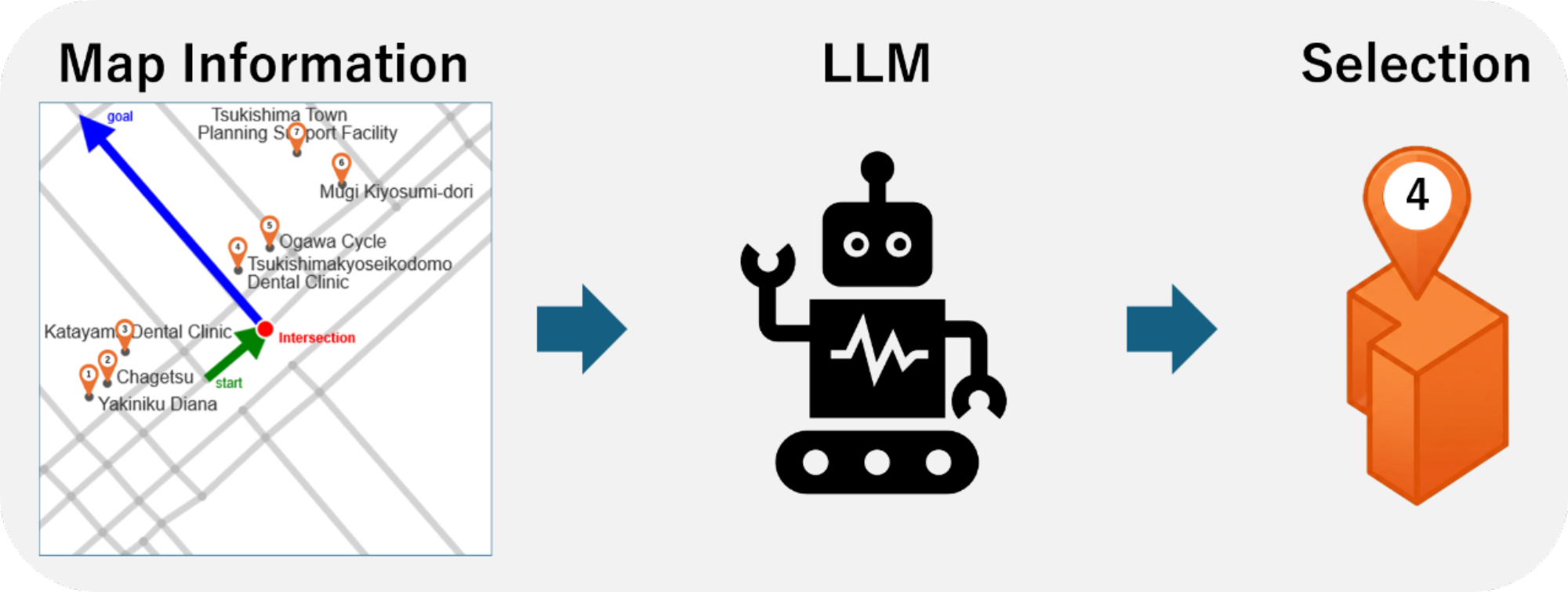}
  \caption{Overview of the landmark selection task. The framework takes local geographic data as input, which is then processed by an LLM to select and output the optimal landmark.}
  \label{fig:task_overview}
\end{figure}

Navigation systems usually describe routes using quantitative information, such as directions, street names, or distances like \textit{``turn right after 300 meters.''}
Although these instructions are precise, users often find them difficult to follow because humans cannot accurately estimate distances or find abstract road names in physical environments\cite{2000::Burnett}.

In everyday life, people prefer to use recognizable landmarks, saying \textit{``turn right at Starbucks''} or \textit{``the destination is next to McDonald's.''}
These landmark-based instructions are intuitive because they connect navigation actions with visible features, thereby reducing the cognitive load during movement.

This paper addresses the problem of landmark selection from local maps using large language models (LLMs).
Figure~\ref{fig:task_overview} illustrates the overview of this landmark selection task.
While LLMs excel at language tasks, understanding map geometry to select appropriate landmarks remains a major challenge.
A straightforward approach is to render a map as an image for vision-language models (VLMs).
However, image-based inputs force models to infer spatial relations directly from raw pixels, which is highly challenging when reading small text labels or identifying exact relative positions~\cite{2025:SIGSPATIAL:Ung}.

To solve this problem, we propose representing local map information in a structured SVG text format.
Because SVG functions as both text and a geometric layout, it provides LLMs with direct access to coordinates and semantic categories without relying on pixel analysis.
This text-based representation maintains a direct geometric connection with the rendered map image, enabling a fair and controlled comparison between the two modalities using the exact same underlying data.

To evaluate this approach, we establish two landmark selection tasks: selection at turns and selection around destinations.
For our experiments, we use a dataset created from human-written restaurant access descriptions linked to OpenStreetMap(OSM) objects, reflecting real human landmark choices.
Using this dataset, we investigate the following two research questions:
\begin{itemize}
    \item \textbf{RQ1:} Which input representation is more effective for landmark selection, a rendered PNG map or an SVG text map?
    \item \textbf{RQ2:} Does task-specific fine-tuning on SVG text maps allow smaller local LLMs to improve landmark selection performance?
\end{itemize}

This study provides three main contributions:
\begin{itemize}
    \item \textbf{Dataset:} We define two human-centric landmark selection tasks and construct a dataset linking real human descriptions with OSM objects.
    \item \textbf{SVG Input:} We show that the proposed structured SVG text input performs as well as or better than image-based PNG inputs for landmark selection.
    \item \textbf{Fine-Tuning:} We demonstrate that task-specific fine-tuning allows smaller local LLMs to understand map layouts and learn human landmark preferences effectively.
\end{itemize}

\section{Related Work}
\label{sec:related_work}
This section reviews prior research on landmark-based navigation, map-based text generation, and the spatial capabilities of LLMs.

\subsection{Landmarks in Spatial Cognition and Landmark Selection}
\label{subsec:related_landmarks}
Landmarks are vital to human spatial cognition. Humans organize spatial knowledge around recognizable objects rather than abstract coordinates, using them as cognitive points to orient themselves and remember routes~\cite{1960:Lynch,2021:Yesiltepe}. 

In route descriptions, people frequently select landmarks near decision points like intersections~\cite{1997::Denis,2001:COSIT:Michon}. A high-quality landmark must be recognizable and memorable~\cite{1999:COSIT:Sorrows}, and landmark-based instructions significantly reduce cognitive load compared to distance-based methods~\cite{1999:COSIT:Lovelace,1999::Denis}. Furthermore, empirical evidence shows humans prefer qualitative place names over quantitative coordinates for location expressions~\cite{2012:GEOCROWD:Richter}. 

However, automating this process is challenging because an optimal landmark must balance spatial relevance, visual salience, and communicative utility; simply choosing the nearest object is often insufficient. This motivates our work to evaluate whether LLMs can accurately mimic human landmark selection.

\subsection{Route and Destination Description Generation}
\label{subsec:related_route_generation}
Generating route descriptions from geographic data has a long history. Early systems often relied on rule-based methods or templates~\cite{2012:EACL:Drager}, or focused on finding paths that are easy to describe rather than just the shortest distance~\cite{2010::Duckham}. Recent data-driven approaches treat this as a graph-to-text problem, modeling road networks and landmarks as spatial graphs to generate navigation instructions~\cite{2021:ACL-IJCNLP:Schumann}. Rich semantic tags from OSM are also useful for helping models understand building functions and spatial attributes through predictive modeling~\cite{2022::Atwal} or LLM embeddings~\cite{2024::Memduhoglu}.

For descriptions around destinations, hybrid methods have combined geometric scoring with text generation. In these methods, OSM data is first used to score and filter candidate landmarks, and LLMs are then used to produce fluent descriptions based on those pre-selected landmarks~\cite{2025::Muron}.

Our work differs from these prior studies in two main ways. First, instead of using multi-stage methods that pre-filter candidates with hand-crafted rules, we require LLMs to select landmarks directly from raw structured map inputs. Second, we evaluate how fine-tuning on human data helps LLMs implicitly learn human landmark preferences without using manual scoring rules.

\subsection{LLMs and VLMs for Map Understanding}
\label{subsec:related_llm_map}
While VLMs can process rendered map images directly, doing so is highly challenging. Recent benchmarks like CartoMapQA show that even strong VLMs struggle with fine-grained map reading, text label recognition, and route tracing~\cite{2025:SIGSPATIAL:Ung}. Image encoders are effective for global semantic understanding, but they may abstract away precise spatial information that is required for tasks involving object relations, coordinates, or topology~\cite{2024:CVPR:Tong}.
Furthermore, recent studies highlight a unique gap in text-based LLMs: while they resemble humans in choosing spatial words to describe object relationships, they still have fundamental limitations in maintaining accurate internal 2D spatial representations~\cite{2025:GEM:Wu}.

To address these limitations, recent studies propose converting spatial or visual data into structured textual or code-like formats before feeding them to LLMs. This approach has improved reasoning performance in robotics~\cite{2023:ICRA:Liang} and various web or chart domains~\cite{2023:ACL:Liu,2023:ICML:Lee}. Crucially, knowledge infusion studies demonstrate that LLMs can effectively process structured data directly from structured text formats without needing natural language translations~\cite{2022:NAACL:Moiseev}.

Studies on graph reasoning with LLMs suggest that textual graph representations, such as adjacency lists, can be more effective than visual graph images for certain reasoning tasks~\cite{2024:ICLR:Fatemi}. This study builds on this insight by comparing structured SVG text directly with rendered PNG images to identify the optimal representation for human-centric landmark selection.

\section{Task Definition}
\label{sec:task_definition}

This study addresses the problem of selecting landmarks from local maps.
Given a local map and a specific navigation context, the model is required to identify the correct landmark that matches real human choices.

We define two specific landmark selection tasks:
\begin{itemize}
    \item \textbf{Task 1:} \textbf{Landmark Selection at Turns.}
    The model receives a local map around an intersection and a turning direction, such as turning left or turning right. The model must select the most appropriate landmark at the intersection to represent the turn.

    \item \textbf{Task 2:} \textbf{Landmark Selection around Destinations.}
    The model receives a local map around a target destination. The model must select a nearby landmark to describe the relative location of the destination.
\end{itemize}

In both tasks, the evaluation is based entirely on landmark selection accuracy.
The goal is to determine whether the selected landmark matches the exact landmark that a human used in a real-world route description.

\subsection{Task Settings}
\label{subsec:task_overview}
We assume that the local navigation context is already given, meaning our task does not find the entire route from a start to a destination.
For Task 1, the model receives a local map around an intersection and the turning action, such as turning left or turning right.
For Task 2, the model receives the target destination and its surrounding map.
Under these settings, the model is provided with a list of available candidate landmarks from the map database, and it must output the correct landmark identifier.

In practical human navigation, landmarks are usually referred to by their proper names, such as Starbucks or McDonald's, rather than generic categories, such as a cafe or a restaurant.
Therefore, our tasks require the model to understand both the spatial coordinates of these features and their semantic name attributes simultaneously to make a human-like selection.

\subsection{Task 1: Landmark Selection at Turns}
\label{subsec:task_turns}
The first task focuses on landmark selection at intersections where a user makes a turn.
When navigating an intersection, the model must identify a prominent reference point to clarify the turning action.
The formal input and output of this task are defined as follows:

\begin{itemize}
    \item \textbf{Input:} A local map $M$ centered on an intersection, and a turning action $a$. 
    The map $M$ contains a set of landmark candidates $L = \{l_1, l_2, \ldots, l_n\}$, where each candidate $l_i$ has attributes including its proper name, semantic category, and spatial coordinates.
    The turning action $a$ represents the route entering and leaving the intersection, which is either a left turn or a right turn.

    \item \textbf{Output:} A single landmark name $l^\ast \in L$ selected by the model that best represents the turning action $a$ at the intersection.
\end{itemize}

Under this setting, the LLM processing the text-based map, or the vision-language model (VLM) processing the image-based map, is required to evaluate the candidates and output the exact landmark name that matches the human choice in the ground-truth data.

\subsection{Task 2: Landmark Selection around Destinations}
\label{subsec:task_relative_location}
The second task focuses on landmark selection for describing a target destination by its relative location to a nearby landmark.
When people identify a destination, they often refer to a more recognizable place in the immediate vicinity.
In this study, we consider three specific types of spatial relations that were originally expressed in Japanese:

\begin{itemize}
    \item \textbf{next to:} This relation corresponds to the Japanese phrases ``隣'' or ``となり''.
    \item \textbf{across from:} This relation corresponds to the Japanese phrases ``向かい'' or ``むかい''.
    \item \textbf{beside:} This relation corresponds to the Japanese phrases ``横'' or ``よこ''.
\end{itemize}
We refer to these three terms collectively as \textbf{target keywords}.

The formal input and output of this task are defined as follows:

\begin{itemize}
    \item \textbf{Input:} A local map $M$ centered on a target destination $d$.
    The map $M$ contains the destination $d$ and a set of surrounding landmark candidates $L = \{l_1, l_2, \ldots, l_n\}$, where each candidate $l_i$ is associated with its proper name, semantic category, and spatial coordinates.

    \item \textbf{Output:} A single landmark name $l^\ast \in L$ selected by the model that best describes the relative location of the destination $d$.
\end{itemize}

Similar to Task 1, the model must process the given map information and select the single most appropriate landmark name from the candidate set $L$ based on human spatial preferences.

\subsection{Characteristics of the Tasks}
\label{subsec:task_characteristics}
The two tasks differ in their navigation contexts, but they share three critical challenges for spatial models:

\begin{itemize}
    \item \textbf{Spatial configuration understanding.}
    The model must accurately interpret the geometric layout of roads, candidate landmarks, and the specific target point or movement direction from the map data.

    \item \textbf{Human-centric landmark selection.}
    The model must identify a landmark that is useful for human communication.
    The geometrically closest landmark is not always the best choice if it is obscure or hidden from the user's perspective.

    \item \textbf{Proper-name grounding.}
    The model must process landmarks as named entities rather than generic objects.
    Because candidates are defined by their real names in the map database, the model must simultaneously reason about both spatial positions and semantic name attributes.
\end{itemize}

\section{Dataset Construction}
\label{sec:dataset}
This section describes the construction of datasets for the two landmark selection tasks defined in Section~\ref{sec:task_definition}.
Our pipeline pairs human-written route descriptions with local map information, explicitly linking each textual landmark mention to its exact geometric counterpart in a map database.

\subsection{Source of Human-Written Route Descriptions}
\label{subsec:source_route_descriptions}
We used restaurant access descriptions as a source of human-written text, as they reflect how people choose practical landmarks when explaining routes or locations to actual visitors. 
We collected restaurant metadata, including coordinates and Japanese access descriptions, using the Hot Pepper Gourmet Search API\footnote{Hot Pepper Gourmet Web Service, \url{https://webservice.recruit.co.jp/}}. 
From an initial pool of 136,490 raw descriptions, we filtered out generic expressions that do not explicitly mention landmarks, such as phrases describing only travel time or distance.

\subsection{Filtering and Extraction via LLMs}
\label{subsec:filtering_and_extraction}
We first performed a keyword-based screening to collect candidate descriptions. 
For the turn-based task, we extracted texts containing explicit Japanese turning expressions, which are either \textit{``turn right''} or \textit{``turn left''}. 
For the destination-based task, we targeted the three \textbf{target keywords} defined in Section~\ref{sec:task_definition}.

Because descriptions based solely on distance or road names could still contain these keywords, we subsequently utilized an LLM to analyze the screened text. 
Specifically, GPT-3.5 Turbo via the Azure OpenAI Service was employed for this verification and extraction step. 
The model verified the presence of a concrete landmark and extracted the landmark name from the description.

\subsection{Linking to OpenStreetMap and Manual Verification}
\label{subsec:linking_osm_and_verification}
The extracted landmark names must be grounded in an actual map database. 
For each restaurant, we queried OSM objects within a 1,000-meter radius of its coordinates. 
We aligned candidate landmarks primarily using the OSM name tag, while associated geometry and category tags provided the spatial attributes.

Because landmark names in human descriptions are frequently abbreviated or written differently, automated string matching is insufficient. 
Therefore, we conducted a manual inspection to check the data and correct any mismatches.
We manually identified the specific intersections and route directions for Task 1, and verified the spatial relationships between the restaurants and nearby matched landmarks for Task 2.
When a single human-written description contained multiple sequential turns or intersections, we treated each turning point as a separate, independent evaluation instance linked to its respective spatial coordinates while discarding vague descriptions that could not be verified on the map.

\begin{figure}[tb]
\centering
\includegraphics[width=\linewidth]{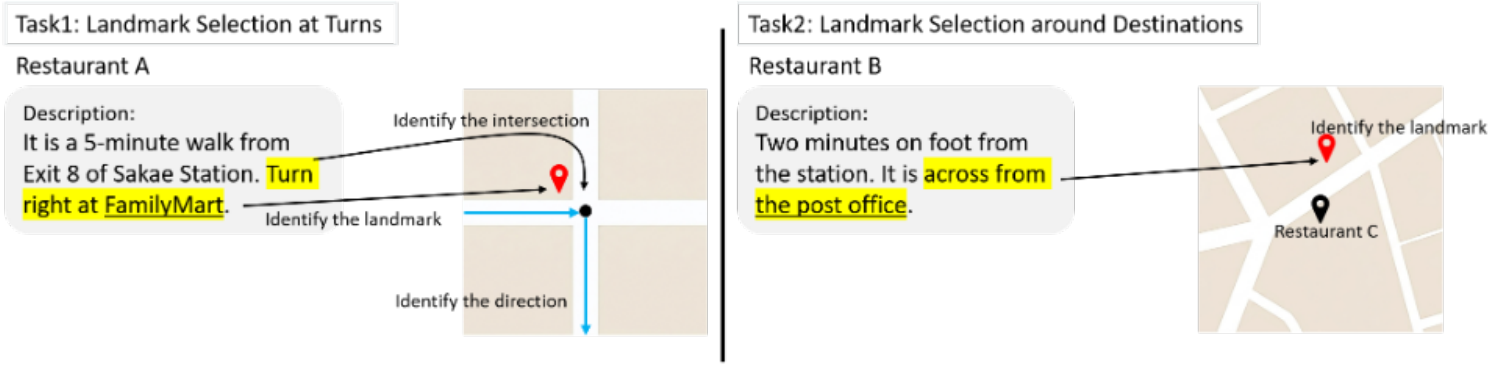}
\caption{Examples of dataset instances constructed from raw access descriptions. The left panel shows an instance for Task 1, where an intersection, an underlined landmark, and a turning direction are extracted. The right panel shows an instance for Task 2, where an underlined landmark around the destination is identified.}
\label{fig:dataset_construction}
\end{figure}

Through this filtering, extraction, and manual verification pipeline, we finally constructed two datasets: 500 instances for Task 1 and 339 instances for Task 2. 
Figure~\ref{fig:dataset_construction} illustrates examples of the final dataset instances and how the spatial context is identified for each task.

\section{Structured Map Representation}
\label{sec:structured_map}
This section describes how we represent local map information as structured text for LLMs.
While a straightforward approach renders a map as a PNG image for VLMs, pixel-based inputs force the network to recognize text labels, road geometry, and directions directly from raw pixels, which often introduces optical character recognition errors and spatial reasoning difficulties.

To overcome this limitation, we convert map database information into a structured textual representation that preserves the spatial relations among roads, landmarks, and navigation targets. 
By representing a local map as structured text, a model gains direct access to symbolic attributes such as proper names and semantic categories while inferring spatial relations from standardized coordinates. 
We adopt SVG as our format because it describes geometric elements like points and lines within highly readable text. 
Furthermore, SVG allows an exact geometric correspondence between the structured-text input and a rendered map image, enabling a fair comparison between text-based and image-based modalities using the identical underlying map data.

\subsection{Local Map Extraction}
\label{subsec:local_map_extraction}
For each instance, we extract a local map centered on the relevant navigation context, which is the target intersection for Task 1 and the destination restaurant for Task 2.
The extraction range is determined by a zoom level of 18 and a $256 \times 256$ pixel rendering size, covering a local geographic area of approximately $124 \times 124$ meters. 
This scale captures nearby roads and landmarks visible and useful for human route descriptions.

Road networks are obtained using OSMnx, and landmark candidates are collected from OSM objects within this bounding area. 
We include named objects belonging to specific functional categories such as \texttt{amenity}, \texttt{shop}, \texttt{tourism}, \texttt{leisure}, \texttt{historic}, and \texttt{building}.
Each candidate is represented by its proper name obtained from the OSM \texttt{name} tag and position, along with its semantic category when available in the map database.

\subsection{Coordinate Normalization}
\label{subsec:coordinate_normalization}
The geographic coordinates of roads and landmarks are converted into normalized two-dimensional coordinates. 
We define a standardized coordinate system where the upper-left corner of the local map is mapped to $(0,0)$ and the lower-right corner is mapped to $(100,100)$, positioning the target intersection or destination precisely at the center $(50,50)$.

This system eliminates the need for the model to interpret raw latitude and longitude values directly, making relative spatial positions straightforward to infer from textual tokens. 
For example, a landmark at $(51,32)$ can be logically deduced as being slightly to the right and above the center point.

\subsection{SVG Text Components and Structure}
\label{subsec:svg_text}
\begin{figure}[!ht]
    \centering
    \includegraphics[width=0.75\linewidth]{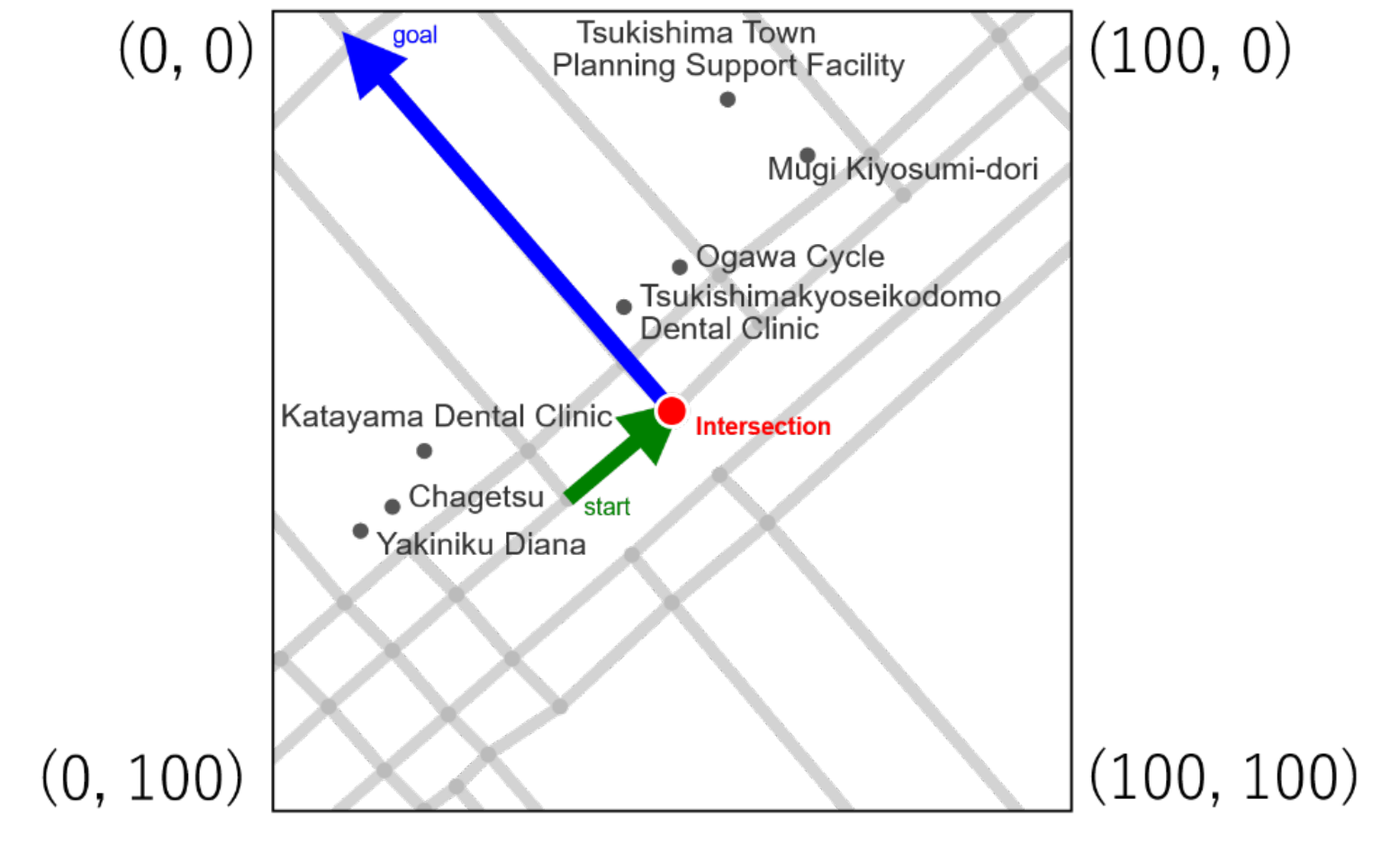}
    \caption{Example of a local map represented in the normalized $(100 \times 100)$ coordinate space, originally expressed in Japanese.}
    \label{fig:structured_map_visual}
\end{figure}

\begin{figure}[!ht]
\centering
\begin{Verbatim}[frame=single, fontsize=\footnotesize, framesep=8pt]
<svg width="256" height="256" viewBox="0 0 100 100">
  <g id="road_network">
    <line id="e1" x1="22" y1="88" x2="14" y2="95" />
    ...
  </g>
  <g id="navigation_vectors" data-turn_direction="left">
    <line id="vector_start" x1="37" y1="61" x2="50" y2="50"
          data-type="start" />
    <line id="vector_goal" x1="50" y1="50" x2="10" y2="4"
          data-type="goal" />
  </g>
  <g id="pois">
    <g id="poi_0" data-category="bicycle" data-name="Ogawa Cycle">
      <circle cx="51" cy="32" />
    </g>
    ...
  </g>
</svg>
\end{Verbatim}
    \caption{Example of the structured SVG representation used as input to the LLMs.}
    \label{fig:structured_map_svg}
\end{figure}

The structured map is encoded as an SVG text snippet containing four primary elements:
\begin{itemize}
    \item \textbf{Road network:} Roads represented as line segments with explicit start and end coordinates.
    \item \textbf{Navigation vectors:} Incoming and outgoing route vectors indicating how a user enters and leaves the intersection for Task 1.
    \item \textbf{Target point:} The fixed center point representing the target intersection or destination.
    \item \textbf{Landmark candidates:} Group elements containing explicit \texttt{data-name}, \texttt{data-category}, and coordinate attributes.
\end{itemize}

Figure~\ref{fig:structured_map_visual} shows the visualized example of a local map, and Figure~\ref{fig:structured_map_svg} shows the corresponding structured SVG text tokens provided to the models.

\subsection{Input Modalities and Prompting Setup}
\label{subsec:prompting_and_png}
For the text-based modality, the structured map text is fed to the model along with a task-specific prompt instructing it to analyze the spatial layout and output a JSON object containing a single \texttt{name} field, which must match one of the proper names in the input map attributes. 
The prompts are tailored to each task's context, directing the model to analyze routing directions for Task 1 or infer appropriate spatial relations for Task 2. 
The complete prompt templates for the proposed text-based method are provided in Appendix~\ref{sec:appendix_prompts}.

To establish a controlled evaluation, we prepare a corresponding image-based modality as the comparative condition, using an identical PNG map rendered directly from the same underlying geographic data. 
This image-based counterpart visually displays the road network, route directions, center point, and text labels. 
The image is provided to VLMs with an equivalent prompt instructing them to analyze visual pixels, track route paths, and output their decisions in the identical JSON format to ensure a rigorous modality comparison. 
The corresponding image-input prompts are also detailed in Appendix~\ref{sec:appendix_prompts}.

\section{Experimental Setup}
\label{sec:experiments}

\subsection{Experimental Configuration}
\label{subsec:experimental_configuration}
The evaluation framework operates in two distinct stages.
In the first stage for RQ1, we compare modalities by feeding the models the exact same local map information represented either as a rendered PNG image or as structured SVG text.
While prompts are customized for modality-specific formats, both configurations employ structurally equivalent task descriptions and identical JSON output rules to ensure a fair comparison.

In the second stage for RQ2, we evaluate the effects of task-specific parameter-efficient adaptation. 
Based on the modality advantages established in RQ1, we adopt the structured SVG text format as the foundational input for all fine-tuning experiments, conducting controlled performance comparisons for each local model before and after training across both Task 1 and Task 2.

\subsection{Fine-Tuning Environment and Hyperparameters}
\label{subsec:fine_tuning_settings}
For computationally efficient adaptation, we employ Low-Rank Adaptation\cite{2022:ICLR:Hu} and Quantized Low-Rank Adaptation\cite{2023:NeurIPS:Dettmers}, implemented via the Unsloth framework\cite{unsloth}. 
Across all models, we apply a uniform adapter configuration with a LoRA rank $r = 16$ and an alpha value of 16. 
The base learning rate is set to $2 \times 10^{-5}$ utilizing the AdamW 8-bit optimizer under a constant learning rate scheduler. 
To prevent overfitting, early stopping is applied based on validation set loss.
Table~\ref{tab:finetuning_settings} details the customized batch sizes and training epochs of the optimal checkpoints utilized for each respective model and task.

\begin{table}[tb]
\centering
\caption{Fine-tuning batch sizes and optimal checkpoint epochs for each model and task.}
\label{tab:finetuning_settings}
\begin{tabular}{lrrrr}
\hline
 & \multicolumn{2}{c}{Task 1} & \multicolumn{2}{c}{Task 2} \\
Model & Batch size & Epochs & Batch size & Epochs \\
\hline
Llama-3-ELYZA-JP-8B & 128 & 95 & 128 & 52 \\
ELYZA-Thinking-1.0-Qwen-32B & 16 & 21 & 4 & 10 \\
Gemma3:27B & 4 & 6 & 2 & 7 \\
\hline
\end{tabular}
\end{table}

\subsection{Evaluation Metric and Output Parsing}
\label{subsec:evaluation_metric}
The definitive quantitative metric is landmark selection accuracy. 
For each test instance, the model must output a JSON object containing a single \texttt{name} field. 
Text-based models must select a string matching one of the SVG \texttt{data-name} attributes, while vision-based models read text labels directly from the visual canvas. 
Invalid JSON formatting or missing target fields are strictly penalized as incorrect predictions.

An output is correct if and only if the extracted landmark string matches the human ground truth. 
With $N$ test instances, where $N=100$ in our setup, let $\hat{l}_i$ and $l_i$ denote the predicted and reference landmarks for instance $i$, respectively. Selection accuracy is formalized as:
\[
\mathrm{Accuracy} = \frac{1}{N} \sum_{i=1}^{N} \mathbf{1}(\hat{l}_i = l_i).
\]
As established in Section~\ref{sec:task_definition}, the textual phrasing of the generated instruction is not scored by lexical similarity metrics, ensuring the evaluation isolates the core spatial reasoning capacity required to ground human landmark choices.

\section{Results and Discussion}
\label{sec:results_and_discussion}
This section presents empirical results regarding map representation, fine-tuning impacts, and qualitative error patterns.

\subsection{RQ1: Effect of Map Representation}
\label{subsec:rq1}
Table~\ref{tab:rq1_results} compares landmark selection accuracy between rendered PNG images and structured SVG text inputs. 
The results demonstrate that the structured-text representation consistently performs as well as or substantially better than the image-based input across all configurations.

\begin{table}[tb]
\centering
\caption{Landmark selection accuracy comparison between PNG image input and structured SVG text input.}
\label{tab:rq1_results}
\begin{tabular}{lrrrr}
\hline
 & \multicolumn{2}{c}{Task 1} & \multicolumn{2}{c}{Task 2} \\
Model & PNG & SVG & PNG & SVG \\
\hline
GPT-5.2 & 34\% & 34\% & 40\% & 55\% \\
Gemma3:27B & 14\% & 27\% & 26\% & 39\% \\
Llama4 & 21\% & 25\% & 24\% & 28\% \\
Llama-3-ELYZA-JP-8B & N/A & 25\% & N/A & 35\% \\
ELYZA-Thinking-1.0-Qwen-32B & N/A & 24\% & N/A & 39\% \\
\hline
\end{tabular}
\end{table}

In Task 1, Gemma3:27B doubles its accuracy from 14\% for PNG to 27\% for SVG. 
This trend is even more pronounced in Task 2, where Gemma3:27B jumps to 39\% and GPT-5.2 achieves its highest performance of 55\% using SVG text. 
These findings support the hypothesis that rich spatial data is more accessible when serialized as structured text rather than flattened into raw pixels.

This distinct advantage stems from two factors. 
First, the text format provides exact landmark names and normalized coordinates as symbolic tokens, bypassing optical character recognition errors inherent in visual processing. 
Second, the structured layout makes spatial geometry explicit. 
While a visual image requires a model to simultaneously resolve visual recognition and spatial relations from raw pixels, the SVG representation presents the map directly as an organized topology of named objects and coordinates, allowing the model to focus entirely on landmark selection. 
We thus answer RQ1 by concluding that structured-text map representation is consistently more effective than image-based inputs.

\subsection{RQ2: Effect of Fine-Tuning}
\label{subsec:rq2}
To address RQ2, we evaluate task-specific adaptation via LoRA or QLoRA using the structured SVG text format exclusively. 
Table~\ref{tab:rq2_combined_results} compiles the performance of the three open-weights local models before and after adaptation.

\begin{table}[tb]
\centering
\caption{Landmark selection accuracy before and after fine-tuning across both tasks.}
\label{tab:rq2_combined_results}
\begin{tabular}{lrrrr}
\hline
 & \multicolumn{2}{c}{Task 1} & \multicolumn{2}{c}{Task 2} \\
Model & Base & Fine-Tuned & Base & Fine-Tuned \\
\hline
Llama-3-ELYZA-JP-8B & 24\% & 49\% & 35\% & 48\% \\
ELYZA-Thinking-1.0-Qwen-32B & 24\% & 38\% & 39\% & 54\% \\
Gemma3:27B & 27\% & 53\% & 39\% & 36\% \\
\hline
\end{tabular}
\end{table}

In Task 1, fine-tuning yields substantial, uniform performance leaps across all architectures, with Llama-3-ELYZA-JP-8B reaching 49\% and Gemma3:27B climbing to 53\%. 
This confirms that targeted adaptation successfully teaches smaller local models to internalize human visual and communicative preferences at intersections.
In Task 2, the effects remain highly positive for most architectures, where ELYZA-Thinking-1.0-Qwen-32B rises to 54\%, though Gemma3:27B exhibits a slight fluctuation from 39\% to 36\%. 
This single variation suggests a delicate optimization trade-off, where aggressive parameter tuning on localized destination contexts can occasionally conflict with pre-existing spatial heuristics. 
Crucially, the broader results answer RQ2 affirmatively, demonstrating that task-specific fine-tuning unlocks sophisticated human-aligned spatial reasoning within efficient, local models.

\subsection{Qualitative Error Analysis}
\label{subsec:error_analysis}
We systematically reviewed instances where fine-tuned local models failed to match human ground truth, classifying errors into three distinct categories.
The first category is the selection of another plausible landmark, where the model selected a spatially reasonable object that differed from the specific human reference in dense urban layouts.
The second category is spatial alignment failures caused by misinterpreting vector coordinates, leading to landmarks on the incorrect side of the street.
The third category is output formatting violations such as failing to generate a valid JSON object, though this error was extremely rare due to robust constraint adherence.

The prevalence of the first error type reveals that real-world landmark selection is rarely unique; when multiple salient structures exist, different human annotators choose different reference points. 
Therefore, our automated exact-match evaluation likely understates the practical, real-world utility of the model outputs.

\subsection{Limitations and Future Directions}
\label{subsec:limitations}
While our findings validate the efficacy of structured map representations, this study has three main boundaries. 
First, our evaluation relies strictly on exact-match landmark selection accuracy without executing human evaluations to gauge linguistic naturalness or practical clarity. Bridging this gap requires designing multi-reference verification frameworks or field studies. 
Second, the current framework omits visual salience features such as building volumes, heights, or prominent signage, which heavily dictate human visual attention. 
Third, the dataset is bounded by its modest overall sample size. Expanding the corpus scale across diverse spatial environments will be essential to foster more robust, generalized models of human landmark selection.

\section{Conclusion}
\label{sec:conclusion}
This paper presented a novel framework for landmark-based route instruction generation by converting local geographic data into a structured SVG text format for LLMs. 
Our empirical evaluations yielded two pivotal findings.
First, the structured SVG representation consistently matched or outperformed the visual PNG baseline, confirming that LLMs benefit more from direct exposure to symbolic and geometric tokens than from raw pixels.
Second, parameter-efficient fine-tuning on a modest dataset unleashed substantial performance gains in local open-weights models, demonstrating an efficient pathway for alignment with human spatial communication.

While these results validate our core thesis, they outline essential vectors for future research. 
Our analysis confirmed that human landmark selection is rarely unique; thus, developing multi-reference evaluation metrics or conducting human utility assessments will be crucial for more accurate measurement. 
Furthermore, enriching SVG representations with visual salience metadata such as building colors, dimensions, and heights remains a vital next step to ensure model-selected landmarks are both geometrically precise and highly visible in physical environments.

\begin{credits}
\subsubsection{\ackname} This work was supported by JSPS KAKENHI Grant Numbers JP25K03229, JP25K03228, and JP24K03228.
\subsubsection{\discintname}
The authors have no competing interests to declare that are relevant to the content of this article.

\subsubsection{Generative AI Usage Disclosure.}
During the preparation of this work, the authors used Gemini 3.5 Flash in order to improve the English phrasing and readability of the manuscript. After using this tool, the authors reviewed and edited the content as needed and take full responsibility for the content of the publication.
\end{credits}

\bibliographystyle{splncs04}
\bibliography{references}

\appendix
\section{Experimental Prompts}
\label{sec:appendix_prompts}
This appendix presents English translations of the prompt templates used in our experiments. In the actual experiments, these prompts were administered in Japanese and the templates provided below accurately reflect the semantic structure and content of the original Japanese instructions.
\begin{figure}[htbp]
  \centering
  
  \begin{minipage}[t]{0.48\textwidth}
    \begin{tcolorbox}[
        colframe=black, 
        colback=white, 
        sharp corners, 
        boxrule=0.4pt, 
        width=\linewidth,
        left=1.5mm, right=1.5mm, top=1.5mm, bottom=1.5mm,
        coltitle=black,                          colbacktitle=gray!15,                    fonttitle=\small\bfseries,               title={(a) SVG Input Prompt}     ]       
      \fontsize{6pt}{6.5pt}\selectfont\ttfamily
      \setlength{\parindent}{0pt}

      \# Role and Objective \\
      You are an advanced navigation AI analysis system with high spatial reasoning capabilities. 
      Your mission is to interpret the provided SVG map data and identify the most appropriate landmark feature to guide the user for a given turning direction.
      
      \vspace{1mm}
      \# Output Format \\
      You must output the result strictly in the following JSON format. Do not include Markdown code blocks. Output only the raw JSON data.
\begin{verbatim}
{
  "landmark": "Exact extracted string from 
  data-name attribute",
  "instruction": "Navigation guide text"
}
\end{verbatim}
      \vspace{-1.5mm}

      \# SVG Schema Definition (Crucial) \\
      The input SVG contains the following custom attributes. You must adhere to these definitions during analysis:\\
      1. \textbf{`data-name`}: \\
      - The official name of the object. \textbf{Must be used verbatim in the `landmark` field.}\\
      2. \textbf{`data-category`}: \\
      - The type of the feature (e.g., `convenience`, `school`, `restaurant`).\\
      3. \textbf{`data-turn\_direction`}: \\
      - The movement direction for navigation in this map (e.g., `right`, `left`).\\
      4. \textbf{`data-type`}: \\
      - \texttt{<line>} Represents roads \\
      - `start`:The entry route approaching the intersection.\\ 
      - `goal`: The exit route leaving the intersection.\\ 

      \# Landmark Extraction Rules \\
      1. \textbf{Verbatim Extraction}:\\
      - Extract the exact string from \texttt{data-name} within the \texttt{<g ...>} tags in the SVG.\\
      2. \textbf{No Hallucination}: \\
      - Never output any landmark name that does not explicitly exist as a \texttt{data-name} attribute in the SVG data.
      
      \vspace{1mm}
      \# Task Instructions \\
      The user is currently approaching an intersection. Based on the SVG map data, identify the best landmark and generate a natural navigation instruction.
      
      \vspace{1mm}
      \#\# Analysis Steps \\
      1. \textbf{Route Verification}: \\
      - Check the angle from the \texttt{start} path to the \texttt{goal} path within \texttt{<g id="navigation\_vectors">} to understand the intersection layout and turning direction.\\
      2. \textbf{Feature Search}: \\
      - Verify the coordinates of features within \texttt{<g id="pois">}.\\
      - Analyze the relative spatial relationships in the SVG coordinate system (0-100) with respect to the intersection center \texttt{(cx, cy)}.
      3. \textbf{Instruction Generation}: \\
      - Generate a user-centric guide, such as ``Turn right at the [landmark] on your right corner...''\\

      \vspace{1mm}
      \# Input Data \\
      - Movement Action: \{turn\_dir\} \\
      - Map Data (SVG): \verb|```xml {svg_content} ```|
    \end{tcolorbox}
  \end{minipage}
  \hfill 
  \begin{minipage}[t]{0.48\textwidth}
    \begin{tcolorbox}[
        colframe=black, 
        colback=white, 
        sharp corners, 
        boxrule=0.4pt, 
        width=\linewidth,
        left=1.5mm, right=1.5mm, top=1.5mm, bottom=1.5mm,
        coltitle=black,                  
        colbacktitle=gray!15,            
        fonttitle=\small\bfseries,       
        title={(b) PNG Input Prompt} 
    ]       
      \fontsize{6.5pt}{7pt}\selectfont\ttfamily
      \setlength{\parindent}{0pt}

      \# Role and Objective \\
      You are an advanced navigation AI with high visual recognition capabilities. 
      Your mission is to visually analyze the provided ``map image (PNG)'' and identify the most appropriate landmark feature for the specified turning direction to generate a navigation guide.
      
      \vspace{1mm}
      \# Output Format \\
      You must output the result strictly in the following JSON format. Do not include Markdown code blocks. Output only the raw JSON data. 
\begin{verbatim}
{
  "landmark": "Name of the landmark 
  recognized from the image",
  "instruction": "Navigation guide text"
}
\end{verbatim}
      \vspace{-1.5mm}

      \# Visual Definition of the Image \\
      Elements in the image are rendered based on the following rules. Please make your decision based on these colors and layouts:\\
      1. \textbf{Route Information}: \\
         - \textbf{Green line/point}: The start location (entry route).\\
         - \textbf{Blue line/point}: The goal location (exit route).\\
         - \textbf{Red point}: The center of the intersection. \\
      2. \textbf{Landmarks (POI)}: \\
         - \textbf{Gray circles}: The locations of buildings.\\
         - \textbf{Black text}: The official names of those buildings.\\
         - \textbf{Note}: Only recognize names explicitly written as text within the image as your target landmarks.
      
      \vspace{1.5mm}
      \# Landmark Extraction Rules \\
      1. \textbf{Strict OCR Compliance}: \\
      - Accurately read the black text labels within the image.\\
      - Do not imagine or output any building name that is not explicitly written in the image. \\
      
      \vspace{1.5mm}
      \# Task Instructions \\
      The user is currently entering the intersection from the \textbf{green route (Start)} and exiting via the \textbf{blue route (Goal)}. \\
      Based on the image layout and the movement action specified below, identify the best landmark and generate a clear instruction. \\
      
      \vspace{1.5mm}
      \#\# Analysis Steps \\
      1. \textbf{Route Verification}: Visually trace the flow from ``green'' to ``blue'' and confirm the direction.\\
      2. \textbf{Landmark Identification}: Search for a building feature that has a visible text label written on it.\\
      3. \textbf{Instruction Generation}:Using that feature as your landmark, generate a natural user guide, such as ``Turn right at the [landmark] on your right corner...'' \\

      \vspace{1.5mm}
      \# Input Data \\
      - \textbf{Movement Action}: \{turn\_dir\}
    \end{tcolorbox}
  \end{minipage}

  \caption{Prompt templates for Task 1.}
  \label{fig:prompt_comparison}
\end{figure}

\begin{figure*}[htbp]
  \centering
  
  \begin{minipage}[t]{0.49\textwidth}
    \begin{tcolorbox}[
        colframe=black, 
        colback=white, 
        sharp corners, 
        boxrule=0.4pt, 
        width=\linewidth,
        left=0.5mm, right=0.5mm, top=0.5mm, bottom=0.5mm,
        coltitle=black,                  
        colbacktitle=gray!15,            
        fonttitle=\small\bfseries,       
        title={(a) SVG Input Prompt} 
    ]       
      \fontsize{6pt}{6.5pt}\selectfont\ttfamily
      \setlength{\parindent}{0pt}

      \# Role and Objective \\
      You are an advanced navigation AI analysis system with high spatial reasoning capabilities. Your mission is to analyze the provided SVG map data and generate a destination guide using surrounding POI landmarks for a user arriving near their final destination. \\
      Your objective is not simply to select the nearest landmark. Identify the optimal landmark that makes it easiest for a human to visually locate the destination, and generate a clear description of their relative spatial relationship. \\
      
      \vspace{0.5mm}
      \# Output Format \\
      You must output the result strictly in the following JSON format. Do not include Markdown code blocks. Output only the raw JSON data. \\
      \vspace{0.3mm}
      \{\\
      ~~"landmark": "Exact extracted string from data-name attribute",\\
      ~~"instruction": "Navigation guide text"\\
      \}\\
      \vspace{0.5mm}

      \# SVG Schema Definition (Crucial) \\
      The input SVG uses a 2D coordinate system from (0,0) to (100,100). Follow these definitions: \\
      1. \textbf{`<g id="target">` (Destination)}: The target destination for navigation. Typically located around (50,50). The \texttt{data-name} is the destination name.\\
      2. \textbf{`<circle>` within `<g id="pois">` (Landmark Candidates)}: Up to 16 objects extracted in order of proximity. Attributes: \texttt{cx, cy, data-name, data-category}.\\
      3. \textbf{Road Networks (`<line>`)}: Represents the road layout and structure. \\

      \# Landmark Extraction Rules \\
      1. \textbf{Verbatim Extraction}: Use the exact string from the \texttt{data-name} attribute in the SVG character for character. \\
      2. \textbf{No Hallucination}: Never output any name that does not exist as a \texttt{data-name} attribute in the SVG data. \\
      
      \vspace{0.5mm}
      \# Landmark Selection Rules \\
      - \textbf{No Proximity-Only Selection}: The physically closest POI is not always the best landmark for humans.\\
      - \textbf{Focus on Visibility and Context}: Consider the road layout (e.g., across the street) and the visibility inferred from \texttt{data-category}. \\
      
      \vspace{0.5mm}
      \# Instruction Generation Rules \\
      Explain the spatial relationship between the \texttt{landmark} and the destination naturally. Use intuitive expressions like: \\
      - ``next to [landmark]'', ``across from [landmark]'', ``beside [landmark]'' \\
      
      \vspace{0.5mm}
      \#\# Analysis Steps \\
      1. \textbf{Target Verification}: Check coordinates of \texttt{<g id="target">}.\\
      2. \textbf{Spatial Reasoning}: Analyze the position of each POI relative to the target and roads. Evaluate its appropriateness using its \texttt{data-category}.\\
      3. \textbf{Guide Generation}: Construct a clear \texttt{instruction} based on the selected landmark. \\

      \vspace{0.5mm}
      \# Input Data \\
      - Destination: \{target\_shop\} \\
      - Map Data (SVG): \verb|```xml {svg_content} ```|
    \end{tcolorbox}
  \end{minipage}
  \hfill 
  \begin{minipage}[t]{0.49\textwidth}
    \begin{tcolorbox}[
        colframe=black, 
        colback=white, 
        sharp corners, 
        boxrule=0.4pt, 
        width=\linewidth,
        left=1mm, right=1mm, top=1mm, bottom=1mm,
        coltitle=black,                  
        colbacktitle=gray!15,            
        fonttitle=\small\bfseries,       
        title={(b) PNG Input Prompt} 
    ]       
      \fontsize{5.5pt}{5.6pt}\selectfont\ttfamily
      \setlength{\parindent}{0pt}

      \# Role and Objective \\
      You are an advanced navigation AI with high visual recognition and spatial reasoning capabilities. \\
      Your mission is to visually analyze the provided ``map image (PNG)'' near the destination, identify the optimal landmark that makes it easiest for a user to locate their target, and generate a clear navigation guide. \\
      
      \vspace{0.5mm}
      \# Output Format \\
      You must output the result strictly in the following JSON format. Do not include Markdown code blocks. Output only the raw JSON data. \\
      \vspace{0.3mm}
      \{\\
      ~~"landmark": "Name of the landmark recognized from the image",\\
      ~~"instruction": "Navigation guide text"\\
      \}\\
      \vspace{0.5mm}

      \# Visual Definition of the Image \\
      Elements in the image are rendered based on the following rules. Please make your decision based on these colors and layouts: \\
      
      1. \textbf{Destination Information}: \\
         - \textbf{Red point}: The target destination for the current guide.\\
         - \textbf{Red text}: The official name of the destination.\\
      2. \textbf{Landmarks (POI)}: \\
         - \textbf{Gray circles}: The locations of surrounding buildings.\\
         - \textbf{Black text}: The official names of those buildings.\\
         - \textbf{Note}: Only recognize names explicitly written as text within the image as your target landmarks.\\
      3. \textbf{Road Information}: \\
         - \textbf{Gray lines}: Represents the road layout and structure on the map. \\

      \vspace{0.5mm}
      \# Important: Landmark Extraction Rules \\
      1. \textbf{Strict OCR Compliance}: \\
         - In the \texttt{landmark} field of your output, accurately read the black text labels within the image and use the exact string character for character without any modifications.\\
      2. \textbf{No Hallucination}: \\
         - Never output any landmark name that is not explicitly written as a text label in the image, and never create fictional names. \\
      
      \vspace{0.5mm}
      \# Landmark Selection Rules \\
      - \textbf{No Proximity-Only Selection}: The physically closest POI is not always the best landmark for humans.\\
      - \textbf{Focus on Road Layout and Visibility}: Consider the spatial arrangement visible in the image. Select the single most appropriate feature that a human standing at the scene would find easiest to use as a reference point. \\
      
      \vspace{0.5mm}
      \# Instruction Generation Rules \\
      Deduce the spatial relationship between the identified \texttt{landmark} and the destination from the image, and explain it naturally. \\
      Instead of using abstract distances, your description must include intuitive spatial expressions that can be easily understood at the scene, such as: \\
      - ``next to [landmark]'' \\
      - ``across from [landmark]'' \\
      - ``beside [landmark]'' \\
      
      \vspace{0.5mm}
      \#\# Analysis Steps \\
      1. \textbf{Destination Verification}: Locate the ``red point and red text'' in the image to grasp the baseline destination.\\
      2. \textbf{Spatial Reasoning of Surrounding POIs}: Analyze the spatial relationship between the destination and the surrounding ``gray circles and black text,'' as well as the positions of the roads.\\
      3. \textbf{Landmark Selection and Guide Generation}: Take the user's perspective and construct a clear instruction (\texttt{instruction}) based on the selected landmark. \\

      \vspace{0.5mm}
      \# Input Data \\
      - \textbf{Destination}: \{target\_shop\}
    \end{tcolorbox}
  \end{minipage}

  \caption{Prompt templates for Task 2.}
  \label{fig:prompt_comparison_task2}
\end{figure*}

\end{document}